\documentclass{article}
\usepackage{spconf,amsmath,graphicx,url,subcaption}
\usepackage{amssymb}
\usepackage{multirow}

\usepackage{todonotes}

\title{An Analysis of Linear Complexity Attention Substitutes with BEST-RQ}
\name{Ryan Whetten$^1$, Titouan Parcollet$^2$, Adel Moumen$^1$, Marco Dinarelli$^3$, Yannick Estève$^1$ 
\thanks{This work received funding from the French ANR E-SSL project (N°ANR-22-CE23-0013) and used HPC resources from GENCI–IDRIS (AD011014732 and A0131013821)}
}
\address{$^1$ Laboratoire Informatique d'Avignon, Avignon Université, France \\
$^2$ Samsung AI Center Cambridge, United Kingdom \\ 
$^3$ Univervisté Grenoble Alpes, Inria, CNRS, Grenoble INP, LIG, 38000, Grenoble, France}

\begin{document}
%
\maketitle
\begin{abstract}
Self-Supervised Learning (SSL) has proven to be effective in various domains, including speech processing. However, SSL is computationally and memory expensive. This is in part due the quadratic complexity of multi-head self-attention (MHSA). 
Alternatives for MHSA have been proposed and used in the speech domain, but have yet to be investigated properly in an SSL setting.
In this work, we study the effects of replacing MHSA with recent state-of-the-art alternatives that have linear complexity, namely, HyperMixing, Fastformer, SummaryMixing, and Mamba. We evaluate these methods by looking at the speed, the amount of VRAM consumed, and the performance on the SSL MP3S benchmark. Results show that these linear alternatives maintain competitive performance compared to MHSA while, on average, decreasing VRAM consumption by around 20\% to 60\% and increasing speed from 7\% to 65\% for input sequences ranging from 20 to 80 seconds.


\end{abstract}
\section{Introduction}
\label{sec:intro}
Self-supervised learning (SSL) is an approach to training machine learning models where pseudo-targets are extracted from the data itself.  
Since SSL is unsupervised, these models can be pre-trained on immense amounts of unlabeled data, and then obtain good results on downstream tasks using minimal amounts of labeled data. 
SSL methods have proven to be useful in a variety of domains including in speech processing~\cite{mohamed2022self}, where SSL has reached state-of-the-art performance in tasks like Automatic Speech Recognition (ASR)~\cite{yang2021superb}, Emotion Recognition (ER)~\cite{yang2021superb}, Automatic Speaker Verification (ASV)~\cite{yang2021superb}, Spoken Language Understanding (SLU)~\cite{yang2021superb}, and Automatic Speech Translation (AST)~\cite{nguyen2020investigating, babu22_interspeech}.

Despite their performance, training SSL models is still very costly in terms of the amount of data, GPUs, and time needed. For example, Google USM was trained on 12 million hours (or over 1,369 years) of audio~\cite{zhang2023google}, and the base and large XLS-R models were trained with 128 and 200 GPUs, respectively~\cite{babu22_interspeech}. Even in efforts to make state-of-the-art models like HuBERT and data2vec more efficient, these SSL models still require around 1,000 A100 GPU hours of training~\cite{chen2023reducing, baevski2023efficient}. 

In a study of the architectures of SSL models for speech~\cite{parcollet2023efficiency}, the authors identified three main culprits: (i) the Acoustic Feature Extractor (AFE), which transforms the raw waveform into a latent representation; (ii) the context encoder, which is often a large Transformer~\cite{vaswani2017attention} or Conformer~\cite{gulati2020conformer}; and (iii) the SSL training objective. 



When looking at these three culprits, BEST-RQ~\cite{chiu2022self} seems to be theoretically one of the most efficient SSL models for speech that has been proposed. BEST-RQ starts with Mel Filterbanks, addressing culprit (i). Then, it creates pseudo labels using a frozen, randomly initialized linear projection and codebook coupled with cross-entropy training, addressing culptrit (iii).
In contrast, other models, such as wav2vec 2.0~\cite{baevski2020wav2vec}, use a learnable codebook and typically use a combination of objectives, slowing down training. For instance, previous studies showed that BEST-RQ obtain comparable downstream performance to wav2vec 2.0 while being 2.4 times faster to train~\cite{whetten2024open}. However, for culprit (ii), the context encoder, BEST-RQ uses Conformer layers, which are computationally expensive.



Conformers, like transformers, are expensive partly due to multi-head self-attention (MHSA) time complexity being quadratic with respect to the input sequence length. Therefore, to address culprit (ii), one needs to search for an efficient alternative to MHSA. Many studies have been conducted to reduce the complexity of MHSA, but only few have been applied to speech tasks and none have have been applied to SSL for speech. Examples of such methods include HyperMixing~\cite{mai23_interspeech}, Fastformer~\cite{wu2021fastformer}, SummaryMixing~\cite{summarymixing}, and Mamba~\cite{gu2023mamba}.

The main contribution of this work is to address culprit~(ii) by, for the first time, evaluating the most promising linear time complexity alternatives to MHSA in an SSL for speech setting. 
Downstream experiments conducted following the MP3S benchmark \cite{zaiem23b_interspeech} show that our linear-time complexity BEST-RQ maintains performance with an equivalent MHSA BEST-RQ while decreasing VRAM consumption by around 20\% to 60\% and increasing inference speed from 7\% to 65\% for input sequences from 20 to 80 seconds. 
As a second contribution, we open-source the code in the widely used SpeechBrain toolkit \cite{ravanelli2021speechbrain}, enabling the community to experiment with efficient SSL models for speech\footnote{\url{https://github.com/whettenr/brq-att-alt-exp}}.

\section{Background} 
\label{sec:background}
Multi-head self-attention (MHSA) has quadratic time complexity with respect to the input length due attention weights being calculated by a dot product between every query-key token pair. Despite its complexity, this dot product operation enables each token to access the global context which is important for reaching high performance~\cite{vaswani2017attention}.


In research on reducing the complexity of MHSA, there are two overarching methods: (i) those aiming to approximate this pair-wise token computation with a lower cost, and (ii) those that do not seek to mimic this pair-wise token computation, but instead introduce global context in another fashion.

Some methods in the first family include sparse attention, such as BigBird and Longformer~\cite{zaheer2020big,Beltagy2020Longformer}, which use a combination of sliding window, global, and random attention to achieve linear complexity. The main issue with these sparse attention methods is that they cannot fully model global context as they operate on windows. Other methods in this family, such as Linformer and Linear Transformer~\cite{wang2020linformer,pmlr-v119-katharopoulos20a} approximate attention by computing low-rank approximates of the key and value matrices or use the dot-product of kernel feature maps and make use of the associative property to achieve linear complexity. Yet, in practice, the aforementioned methods are still computationally expensive~\cite{wu2021fastformer}. 


In contrast, Fastformer~\cite{wu2021fastformer} has proven to perform well on text data, outperforming previously mentioned sparse attention methods and low-rank approximates. This is done by making use of additive attention and element-wise multiplication to summarize the query and key matrices as vectors resulting in linear complexity while fully modeling global context. Fastformer has also been applied to speech and proved to work well on ASR tasks with the branchformer architecture~\cite{peng2022branchformer}. Due to these performances and the availability of its implementation, Fastformer is considered as a good representative candidate of the first family of methods.

The other family of solutions, which do not aim to approximate the self-attention mechanism, include, HyperMixing~\cite{mai2023hypermixer}, SummaryMixing~\cite{summarymixing}, and Mamba~\cite{gu2023mamba}. HyperMixing was first introduced with the HyperMixer~\cite{mai2023hypermixer}. The HyperMixer is an extension of the MLPMixer~\cite{tolstikhin2021mlp}. The disadvantage of MLPMixer is that it can not handle inputs of varying length, making it not suitable for NLP or speech tasks. HyperMixer extends MLPMixer by using hypernetworks~\cite{ha2016hypernetworks} to capture global context for inputs that vary in length. HyperMixer proved to have good performance on text tasks and fully-supervised ASR with the HyperConformer~\cite{mai2023hypermixer,mai23_interspeech}. 

Alternatively, SummaryMixing does not use hypernetworks, but instead the input is passed through a parametrized function, such as a multi layer perceptron, which is averaged across all time steps, resulting in a summary vector. To introduce global context to the input sequence, this summary vector is concatenated to the input at each time step and then fed through another parametrized function, which becomes the final output. SummaryMixing proved to perform just as well as MHSA on ASR, keyword spotting, and SLU, while decreasing the amount of required memory and training time.  

Lastly, Mamba is a method for sequence modeling based on state space models~\cite{gu2023mamba}. State space models can be thought of as a combination of recurrent neural networks and convolutional neural networks, which scale linearly with respect to the sequence length. Mamba has shown good performance on text, audio (which were in an auto regressive SSL setting), and a two supervised speech tasks~\cite{zhang2024mamba}: ASR and speech enhancement. 
HyperMixing, SummaryMixing, and Mamba all have available implementations, and are therefore selected as representatives of the second family of methods.

Despite the good performance of these substitutes for MHSA on supervised tasks, their performance in an SSL setting for speech tasks has not yet been studied, and the focus of these studies has been mostly on ASR. Furthermore, these methods have not been scaled to large models, e.g. from only about 5 to 100M parameters for speech tasks~\cite{mai23_interspeech, summarymixing, peng2022branchformer,zhang2024mamba}, which contrasts with the current scale of SSL models. Thus, in this work we explore these unexamined aspects of linear complexity alternatives by (i) experimenting in an SSL setting, (ii) looking at performance on a variety of speech tasks using a benchmark from the community, and (iii) scaling up model size to above 300M parameters.


\section{Attention Alternatives in BEST-RQ}
\label{sec:att-alt}

In this section, we give an overview of BEST-RQ and describe how Fastfastformer, SummaryMixing, HyperMixing, and Mamba achieve linear complexity. Let us define the input to the self-attention module as $\mathbf{X} \in \mathbb{R}^{T \times d} = [\mathbf{x}_1, \mathbf{x}_2, ...,\mathbf{x}_T]$ or a sequence of vectors of dimension $d$ with $T$ time steps. The difference between methods lies in how this input is transformed to contain information about the global context. \\

\noindent\textbf{BEST-RQ} is, to this day, the most efficient SSL paradigm. It is notably used by large scale models such as Google USM~\cite{zhang2023google} and Universal-1 from AssemblyAI. BEST-RQ begins with Mel Filterbanks which are passed through two paths, (i) the model, and (ii) the random-projection quantizer. For path (i), a random portion of the Mel Filterbanks are masked and then passed through two convolutional layers, a series of conformer layers, and then a linear layer. For path (ii) the unmasked Mel Filterbanks are passed through a randomly initialized frozen linear layer. Then a codebook look up is performed following:

\begin{equation}
y = \arg\min_{i} \left\| norm_{l2}({c}_i) - norm_{l2}(Am) \right\|,
\end{equation}

where $m$ is a stacked portion of four Mel Filterbank frames, $A$ is the linear projection, and $c$ is the codebook. The index, $y$, from this codebook look up is used as the pseudo-target for pre-training for that portion of the input. Then, the cross-entropy loss is calculated between the masked sections of the output of the linear layer from the path (i) and the corresponding pseudo-targets from path (ii). BEST-RQ originally uses self-attention and, therefore, exhibits quadratic time complexity.\\


\noindent\textbf{Fastformer.} The complexity is reduced to linear by using additive attention to summarize the attention matrices as a single vector. To show how this is done, let $\mathbf{Q},  \mathbf{K} \in \mathbb{R}^{T \times d}$, be the standard linear transformations from the transformer composed of $\mathbf{Q} = [\mathbf{q}_1, \mathbf{q}_2, ...,\mathbf{q}_T]$ and $\mathbf{K} = [\mathbf{k}_1, \mathbf{k}_2, ...,\mathbf{k}_T]$. $\mathbf{Q}$ is summarized as a query vector $\mathbf{q}$ by the following:

\begin{equation}
 \alpha_t = \text{softmax}(\frac{\mathbf{w}_q^T \mathbf{q}_t}{\sqrt{d}});
 \quad
 \mathbf{q} = \sum^T_{t=1}\alpha_t \mathbf{q}_t.
\end{equation}

where $\mathbf{w}_q \in \mathbb{R}^d$ is a learnable vector used with each vector of $\mathbf{Q}$ to generate the attentions score $\alpha_t$. The summary query vector $\mathbf{q}$ is calculated as a weighted some of the attention scores with their respective $\mathbf{q}_t$. Then, $\mathbf{q}$ is multiplied by each vector in $\mathbf{K}$, resulting in a global context-aware key matrix that relies on the more efficient element-wise multiplication instead of dot products. \\ 



\noindent\textbf{HyperMixing} reduces the complexity by using a token mixing multi-layered perceptron (TM-MLP) from the MLP-Mixer~\cite{summarymixing}, which can be described as:

\begin{equation}
\mathbf{TM\text{-}MLP}(\mathbf{X}) = \mathbf{LayerNorm}(\mathbf{W}_1(\sigma(\mathbf{W}^T_2 \mathbf{X}^T))),
\end{equation}

where $W_1$, $W_2$ are weight matrices and $\sigma$ represents some non-linear activation function. This is a standard MLP except the linear operations are performed on the transposed $\mathbf{X}$. This can be taught of as operating on the tokens instead of the channels which introduces global context as it allows information to pass between tokens. The key difference between MLP-Mixer and HyperMixing is that $\mathbf{W}_1$and $\mathbf{W}_2$ are generated by another MLP and thus can vary in length. \\

\noindent\textbf{SummaryMixing}. The input, $\mathbf{X}$, is passed into two functions $f$ and $s$ taking the form, in practice, of two MLPs. The output of $s$ is averaged across all time steps, resulting in a summary vector $\bar{s}$. To introduce global context to the input sequence, $\bar{s}$ is concatenated to the output of $f$ at each time step, $t$, and then fed through another function, or MLP in this case, called $c$. The output of $c$ becomes the final output $\mathbf{h}$. The overall SummaryMixing mechanism can be expressed as:

\begin{equation}
\bar{s} = \frac{1}{T} \sum_{t=1}^{T} s(\mathbf{x}_t);
\quad
\mathbf{h} = c(f(\mathbf{X}), \bar{s}).
\end{equation}

Calculating the summary vector $\bar{s}$ is an average and thus is linear with respect to the input length. As a result, SummaryMixing is able to introduce global context with linear complexity.


\noindent\textbf{Mamba.} The memory complexity is reduced to linear by processing the input sequence in an unidirectional manner similar to a recurrent network. Being a selective state space model, Mamba represents the past information of $\mathbf{X}$ as an hidden state $\bold{h}_t$ with constant memory consumption, hence solving the problematic quadratic time complexity from MHSA. To do so, the model employs a discretized state transition matrix $\bar{A} \in \mathbb{R}^{N \times N}$, an input discretized projection matrix $\bar{B} \in \mathbb{R}^{H \times 1}$, and an output projection matrix $C \in \mathbb{R}^{1 \times H}$ to compute the hidden state $\bold{h}_t$ as follow:

\begin{equation}
\label{eq:mamba_1}
 \bold{h}_t = \bar{A}\bold{h}_{t-1} + \bar{B}\bold{X}_t, 
 \quad
  y_t = C\bold{h}_{t}.
\end{equation}
 
 To reduce the   computation overhead brought by the recurrence, an hardware-aware parallel scan algorithm introduced in \cite{gu2023mamba} allows to unroll equation~\ref{eq:mamba_1} as the input sequence $\mathbf{X}$ convolved with a structured kernel composed of $\bar{A}, \bar{B}$, and $C$ fixed. With this improvement, the throughput of Mamba is five times higher than a Transformer. Finally, to introduce global context to the model, similar to MHSA, we use a bidirectional Mamba as motivated by \cite{zhang2024mamba}.


\begin{figure*}[!ht]
    \centering
    \includegraphics[width=1.0\linewidth]{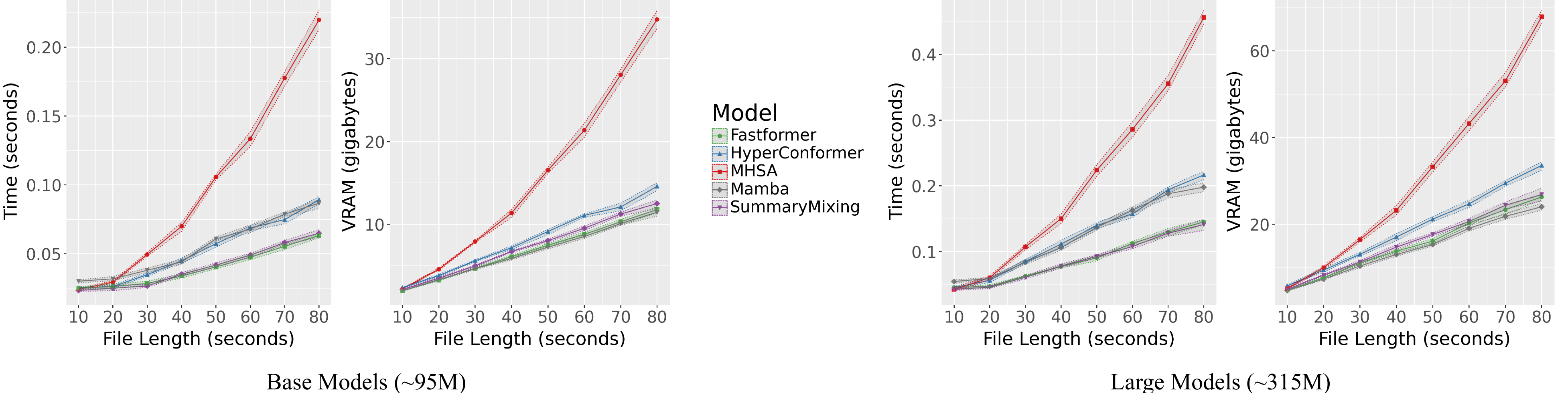}
    \caption{ Inference speed and peak memory of BEST-RQ base and large models with various types of attention. Input length is increased from 10 to 80 seconds. Multi-head self-attention (MHSA) requires significantly more time and VRAM as input size increases where as the alternatives do not.}
    
  \label{fig:speed-mem}
\end{figure*}

\section{Experiments}
\label{sec:methods}
In this section, we describe our SSL pre-training protocol and summarize the downstream tasks, giving a brief overview of the datasets and evaluation metrics for each task.

\subsection{Self-supervised Pre-training}

All models are open-sourced in the SpeechBrain~\cite{ravanelli2021speechbrain} library along with the hyperparameters for each model. \\

\noindent\textbf{Architecture details.} We use the implementation of BEST-RQ from~\cite{whetten2024open} which uses Conformer layers~\cite{gulati2020conformer} with relative sinusoidal positional embedding and the PyTorch multi-head self-attention. For all models the pre-training framework stays the same. We only replace the MHSA cell with either Hypermixing, Fastformer, SummaryMixing, or Mamba. The small models have 12 conformer layers and the large models have 24 layers. 
 We adjust the hidden dimensions to make all the models have around the same number of parameters, that is, 95M and 315M for the small and large ones respectively. The detailed list of hyperparameters can be found in the open-sourced SpeechBrain recipes. \\

\noindent\textbf{Pre-training details.} We pre-train all our models for the same amount of steps, set to 200k using the 960 hours of training data from the LibriSpeech~\cite{panayotov2015librispeech} dataset. Although 200k steps is a lower number than other state-of-the-art models trained for 400k or 800k steps~\cite{baevski2020wav2vec,chen2023reducing}, previous research has empirically shown that,  200k steps is sufficient to compare SSL model performance~\cite{parcollet2023efficiency, whetten2024open}. Dynamic batching is used, meaning audio files are grouped or bucketed together with those of a similar length, and the number of samples per batch varies based on the bucket size in order to keep the number of seconds of input per batch similar. In our experiments, and following the literature~\cite{baevski2020wav2vec}, we set the batch size to 1.7 hours for all models.

    

\begin{table*}[!ht]
    \centering
    \begin{small} 
    \caption{
        Results of various alternatives to MHSA for speech SSL models on the MP3S benchmark. Base models are set to have around 94M parameters and large models, denoted -LG, around 315M. Linear complexity alternatives perform competitively with multi-head self-attention (MHSA) even surpassing depending on the task and probe.
    }
    \label{tab:mp3-results}

    \begin{tabular}{@{\hskip 2pt}c@{\hskip 0pt}c@{\hskip 2pt}c@{\hskip 8pt}c@{\hskip 4pt}c@{\hskip 4pt}c@{\hskip 4pt}cccc@{\hskip 3pt}c@{\hskip 3pt}c@{\hskip 2pt}}

        \hline
        \textbf{Model/Task}
        & \multicolumn{2}{c}{\textbf{Specs}}
        & \multicolumn{4}{c}{\textbf{LibriSpeech train-100 ASR}} 
        & \multicolumn{2}{c}{\textbf{CommonVoice}}
        & \textbf{VoxCeleb}
        & \textbf{SLURP} 
        & \textbf{IEMOCAP} \\
        
        Metric &&& \multicolumn{4}{c}{WER $\downarrow$} 
        & WER $\downarrow$
        & WER $\downarrow$
        & EER $\downarrow$
        & Acc. $\uparrow$
        & Acc. $\uparrow$\\
        \hline

        1st Probe &  & & \multicolumn{4}{c}{LSTM} & Lin. & Lin. & Xvectors & Lin. & Lin. \\
        \hline
        & \multicolumn{2}{c}{\# Params.} &               Clean & Clean LM & Other & Other LM & Welsh & Bask &  ASV &  IC  & ER   \\
        \hline
        MHSA           & \multicolumn{2}{c}{94.0M} & \textbf{10.31} &   \textbf{6.88}   & \textbf{24.95} &  \textbf{18.54}   & 80.04 & 82.56 &  9.36 & 50.3 & 60.1 \\
        HyperConf      & \multicolumn{2}{c}{94.3M} & 10.96 &   7.24   & 25.82 &  19.05   & \textbf{79.65} & \textbf{81.41} &  \textbf{8.54} & 53.2 & 60.1 \\
        Fastformer     & \multicolumn{2}{c}{93.9M} & 13.21 &   8.59   & 33.11 &  24.24   & 82.03 & 84.62 & 12.27 & 36.2 & 56.4 \\
        SummaryMix     & \multicolumn{2}{c}{94.1M} & 11.92 &   7.76   & 28.25 &  20.80   & 79.97 & 81.99 & 11.13 & 52.1 & 63.0 \\
        Mamba          & \multicolumn{2}{c}{94.2M} & 10.46 & 6.90 & 26.00 & 18.88 &  80.71 & 83.66 & 11.91 & \textbf{53.5} & \textbf{64.10} \\
        \hline
        MHSA-LG        & \multicolumn{2}{c}{313.5M} &  \textbf{7.10} &   \textbf{4.98}  & \textbf{17.63} &  \textbf{12.92}  & 84.46 & 88.88 &  8.60 & 54.8 & 62.1 \\
        HC-LG          & \multicolumn{2}{c}{315.4M} &  7.73 &   5.47  & 18.39 &  13.83   & 77.66 & 79.94 &  8.64 & 62.3 & 62.8 \\
        Fastformer-LG  & \multicolumn{2}{c}{315.9M} & 18.74 &  11.83  & 41.07 &  30.55   & 82.85 & 84.59 & 10.13 & 34.5 & 56.9 \\
        SummaryMix-LG  & \multicolumn{2}{c}{313.7M} &  7.22 &   5.04  & 17.67 &  13.14   & \textbf{76.94} & \textbf{78.88} &  \textbf{8.30} & \textbf{62.8} & 65.3 \\
        Mamba-LG       & \multicolumn{2}{c}{314.9M} & 7.84 & 5.23 & 20.57 & 14.66 & 80.86 & 84.66 & 10.54 & 57.7 & \textbf{66.15} \\

        \hline
        \hline
        2nd Probe &  & & \multicolumn{4}{c}{Contextnet} & LSTM & LSTM & ECAPA & LSTM+Lin. & ECAPA \\
        \hline
                       & Time(h) & Mem(GB) & Clean & Clean LM & Other & Other LM & Welsh & Bask &  ASV &  IC  & ER   \\
        \hline
        MHSA           & 453 & 8.86 & \textbf{10.10} &    6.30  & \textbf{22.84} &  16.83   & 57.78 & 50.24 & 4.37 & 74.9 & 60.83 \\
        HyperConf      & 438 & 8.60 & 10.66 &    6.78  & 23.46 &  17.54   & 56.74 & 48.39 & \textbf{3.65} & \textbf{76.9} & 60.10 \\
        Fastformer     & 357 & 7.25 & 13.10 &   8.10   & 29.98 &  22.75   & 59.14 & 52.59 & 4.84 & 68.3 & 54.30 \\
        SummaryMix     & 368 & 7.68 & 11.55 &    7.21  & 25.13 &  18.72   & 58.24 & 50.05 & 4.09 & 76.0 & \textbf{64.34} \\
        Mamba          & 427 & 6.62 & 10.27 & \textbf{5.96} & 23.93 & \textbf{16.37} & \textbf{55.22} & \textbf{47.51} & 4.51  & 76.2 & 63.05 \\

        \hline
        MHSA-LG        &  973 & 20.70 &  7.02 &   4.40 & 15.78 &  \textbf{11.82} & 58.39 & 50.35 & 3.96 & 74.3 & 63.74 \\
        HyperConf-LG   & 1033 & 22.28 &  7.62 &   4.86 & 16.79 &  12.48    & 55.75 & 47.99 & \textbf{3.45} & 78.0 & 64.62 \\
        Fastformer-LG  &  780 & 18.76 & 18.41 &  10.97 & 38.64 &  28.92    & 61.18 & 54.49 & 4.26 & 68.6 & 55.44 \\
        SummaryMix-LG  &  805 & 19.49 &  \textbf{6.79} &   \textbf{4.38} & \textbf{15.69} &  11.83 & 58.24 & 48.59 & 4.05 & \textbf{78.6} & \textbf{65.21} \\
        Mamba-LG       & 1010  & 16.99 & 7.54 & 4.51 & 18.49 & 12.63 & \textbf{55.42} & \textbf{47.59} & 3.76 & 78.1 & 64.21 \\
        \hline
    \end{tabular}
    \end{small}

\end{table*}

\subsection{Downstream Tasks and Metrics}
For the downstream evaluation we use a portion of the tasks from the \emph{Multi-Probe Speech Self-Supervision} benchmark or MP3S~\cite{zaiem23b_interspeech}. 
In MP3S, pre-trained models are frozen and a learned weighted sum of the outputs of the hidden layers of the pre-trained models are fed into a downstream model called a \emph{probe}. Because results were shown to vary depending on the probe’s architecture, for each task two probes are provided. Furthermore, and as SSL models are commonly fine-tuned with the downstream use case, we added an evaluation with a full fine-tuning for ASR.\\

\noindent\textbf{Automatic Speaker Verification (ASV)} is a binary classification task where given 2 utterances, the goal is to determine whether the speakers are the same. The evaluation metric for ASV is the Equal Error Rate (EER). For the dataset, we use VoxCeleb1~\cite{nagrani2017voxceleb}, which is made of utterances from celebrities sourced from YouTube. The dataset is divided into train and test splits, which we used accordingly. The probes for this task are the X-Vectors~\cite{8461375} and ECAPA-TDNN~\cite{desplanques20_interspeech}.\\ 

\noindent\textbf{Intent Classification (IC)} is a classification task of predicting the main purpose or objective of an utterance. We use accuracy as the evaluation metric. For this task, we use the SLURP dataset~\cite{bastianelli-etal-2020-slurp} containing 18 intents coming from single-turn user interaction with a voice assistant. Some examples of intents are \emph{email}, \emph{calendar}, or \emph{play} (as in \emph{play a song}). The probes for this task are a linear probe, in which the output of the SSL model are average-pooled along the time dimension and then passed into a linear classifier, and an LSTM probe, which consists of a two-layered BiLSTM followed by a linear classification layer.\\  

\noindent\textbf{Emotion Recognition (ER)} is the task of predicting the emotion of a speaker. Similar to IC, we use accuracy to measure performance. We use the IEMOCAP dataset, which consists of 10 actors performing scripts with four different emotions (neutral, happy, sad and angry). The probes for this task are a linear probe, like IC, and an ECAPA-DNN probe, like ASV.\\ 

\noindent\textbf{Automatic Speech Recognition (ASR)} is the task of transcribing what was said in an utterance. For ASR, we measure performance by the word error rate (WER). We use the LibriSpeech \emph{train-clean-100} for training, \emph{dev-clean} for validation, and \emph{test-other} for final testing. We report on the final test WER without a language model and with the official 4-gram language model\footnote{\url{openslr.org/11/}} using beam search and shallow fusion. Also, and to test performance in low-resource and out-of-domain settings, which is one use-case of SSL models, we use the two low-resource language ASR datasets proposed in MP3S, Welsh and Basque from the CommonVoice~11.0 dataset~\cite{ardila-etal-2020-common}. For LibriSpeech, we use the 2-layered BiLSTM (LSTM) and ContextNet~\cite{han20_interspeech} probes, and for CommonVoice we use the linear and LSTM probes offered by MP3S. As mentioned, it is common for pre-trained models to be fine-tuned on a given task instead of being frozen. As a representative of this, we fine tune the models using the 100 hours of labeled data in the \emph{train-clean-100} split of LibriSpeech and evaluate on the \emph{test-clean} and \emph{test-other} splits. The downstream architecture for this task is a feed-forward neural network with CTC loss.

\section{Results}
\label{sec:results}

We first evaluate the speed and memory gains on a controlled toy task (Figure~\ref{fig:speed-mem}). The findings are then extended to real SSL pre-training and the MP3S benchmark. \\

\noindent\textbf{Speed and Memory Evaluation}. We give the models randomly generated data of lengths that range from 10 to 80 seconds with 10 seconds intervals. We set the batch size to 6, which was chosen to prevent running out of memory too quickly with MHSA. We perform 10 runs at each input length and time the forward pass as well as measure the VRAM consumption. Only the forward pass is evaluated to concur with a deployment scenario. Averages of computation time and peak VRAM over the 10 runs alongside the 95\% bootstrapped confidence interval are plotted in Figure~\ref{fig:speed-mem}. Measurements were taken on an isolated node with a Nvidia A100 80GB GPU.  

For input sequences of 10 seconds, the amount of time and VRAM needed is relatively similar for all models. However, a difference starts to appear with 20 seconds, where the alternatives, on average, use 24\% less memory, and run 7\% faster relative to MHSA. On the other extreme, at an input of 80 seconds, the alternatives use on average 64\% less memory and run 65\% faster. SummaryMixing and Fastformer were the fastest compared to MHSA with 15\% and 11\% increase in speed with an input of 20 seconds and 70\% and 71\% increase in speed at 80 seconds, respectively. In terms of memory, Mamba and Fastformer proved to be the most memory efficient with 28\% and 29\% decrease in peak memory with an input of 20 seconds, and 67\% and 66\% decrease in memory at 80 seconds, respectively. These findings, however, need to be validated with real-scale experiments to make sure that these alternatives can reach decent downstream performance.

For pre-training time estimates, we run 5 epochs and measure the time and max VRAM on Nvidia V100 32GB GPUs keeping the batch size at 1.7 hours as in the full pre-training. We scale these numbers up to 200k steps to get an estimate of the full number GPU hours. All of the alternatives prove to be faster and use less memory except the HyperMixing-LG and Mamba-LG models. We report these in Table~\ref{tab:mp3-results}. \\





\setlength{\tabcolsep}{2pt}
\begin{table}[!h]
    \centering
    \begin{tabular}{ccccc}
        \hline
        \textbf{Model}
        & \multicolumn{4}{c}{\textbf{Fine-Tune LibriSpeech train-100}}  \\
        &          Clean & Clean LM (C.I.) & Other & Other LM \\
        \hline
        BRQ    &  \textbf{7.01} & \textbf{5.35} ($\pm$ 0.26) & \textbf{16.98} & \textbf{13.55} \\
        HyperConf     &  8.22 & 5.77 ($\pm$ 0.28) & 19.29 & 15.03 \\
        Fastformer   & 9.32 & 6.82 ($\pm$ 0.31) & 22.75 & 17.95 \\
        SummaryMix    &  8.72 & 6.30 ($\pm$ 0.28) & 21.78 & 16.97 \\
        Mamba    &  7.61 & 5.50 ($\pm$ 0.28) & 19.97 & 15.37 \\
        \hline
        BRQ-LG &  \textbf{5.03} & \textbf{3.98} ($\pm$ 0.21) & \textbf{11.52} &  \textbf{9.42} \\ 
        HyperConf-LG  &  5.87 & 4.54 ($\pm$ 0.32) & 13.13 & 10.78 \\
        Fastformer-LG  & 13.16 & 9.89 ($\pm$ 0.34) & 31.91 & 26.75 \\
        SummaryMix-LG &  5.28 & 4.20 ($\pm$ 0.25) & 12.80 & 10.50 \\
        Mamba-LG & 5.59 & 4.48 ($\pm$ 0.25) & 15.47 & 12.66 \\
        \hline
    \end{tabular}
    \caption{Results with fine-tuning on LibriSpeech~\emph{train-100}. For the \emph{test-clean} set, we report th confidence interval (C.I.).}
    \label{tab:ft-results}
\end{table}

\noindent\textbf{Speech Recognition on LibriSpeech}. With the LSTM probe (Table \ref{tab:mp3-results}), MHSA performed better than Mamba, HyperMixing, and SummaryMixing. For the base models with the Contextnet probe, the base version of Mamba outperformed MHSA. For larger models with the Contextnet probe, SummaryMixing performed better than MHSA except with the LM on the \emph{test-other} split, in which SummaryMixing was 0.01 behind MHSA. Despite a dedicated parameter tuning, the large Fastformer falls significantly behind the others.

We then fine-tuned the SSL models on the \emph{train-100} subset of LibriSpeech (Table~\ref{tab:ft-results}). Confidence intervals come from 1000 bootstraps on the \emph{test-clean} set with a LM. Interestingly, the confidence intervals of HyperMixing and Mamba base models and the SummaryMixing Large model overlap with MHSA best performance. It is therefore unclear if MHSA would always outperform these alternatives. \\




\noindent\textbf{Speech Recognition on CommonVoice}. This dataset tells a different story and we hypothesis that this is due to a major issue with most speech SSL model benchmarks. Indeed, Librispeech is always used both during the pre-training and downstream evaluation, introducing a clear bias. With CommonVoice (Table \ref{tab:mp3-results}), and with both probes, MHSA is not the best model anymore. With the linear probe, HyperConformer and SummaryMixing both performed better than MHSA on Welsh and Bask. With the LSTM probe, Mamba performed the best followed by HyperMixing, then SummaryMixing or MHSA depending on the language. This demonstrates the ability of these alternative methods to generalize well to out-of-domain and low-resource datasets compared to MHSA. \\


\noindent\textbf{Speaker Verification}. As for CommonVoice, results varied depending on the probe used, however, MHSA clearly is not the best performing solution (Table \ref{tab:mp3-results}). For the ECAPA probe HyperMixing proved to be the best with an EER of 3.65 and 3.54 for the base and large models respectively. With the Xvectors probe, HyperMixing performed best out of the small models with an EER of 8.54 and for the large model SummaryMixing performed best with and EER of 8.30. \\

\noindent\textbf{Intent Classification}. The accuracies observed on this task validate our previous findings (Table \ref{tab:mp3-results}). All alternatives, except Fastformer, and with both probes performed significantly better than MHSA. \\




\noindent\textbf{Emotion Recognition}. This task tells a similar story to intent classification, but with a slightly different leader board (Table~\ref{tab:mp3-results}). Again, MHSA is not the best performing solution. Indeed, Mamba performed the best with the linear probe reaching an accuracy of 64.10\% and 66.15\% for the base and large model respectively, compared to 60.1\% and 62.1\% for MHSA. For the ECAPA-TDNN probe, SummaryMixing performed the best with and accuracy of 64.34\% and 65.21\% for the base and large model respectively. \\

\section{Discussion and Conclusion}
\label{sec:dis-con}
One of the goals of SSL pre-training is to develop a model that can represent data without the need of labeled data in a way that is useful for a variety of downstream tasks. 
As this process is expensive, making SSL speech models less resource intensive is an active area of research. Part of this expense is due to multi-head self-attention. While alternatives exist, when it comes to speech tasks, they have only been applied in fully supervised tasks. In this work, we explore replacing multi-head self-attention with four state-of-the-art alternatives: HyperMixing, Fastformer, SummaryMixing and Mamba in a SSL setting for speech tasks. We show that for sequences 20 seconds or more these alternatives are substantially faster and consume less memory. 
Based on the toy test, and considering that about 75\% of LibriSpeech is below 15~seconds, we believe that would see a greater difference in the pre-training time and memory when pre-training with a set such as \cite{chiu2022self} where all pre-training audio was cropped to be between 32 and 64 seconds.

However, for small input under approximately 20 seconds the amount of time and memory a model consumes are all relatively similar. This threshold, however, could reduce drastically if the model does not combine Mel Filterbanks with a two-dimensional CNN, as the acoustic feature extractor plays a critical role in the length of the sequence arriving to the self-attention module. For instance, one may expect that replacing this CNN with a one-dimensional one would bring this threshold to 10 seconds. Nevertheless, many files in common speech datasets are shorter than 20 seconds and many speech tasks do not require global context from anything over 20 seconds. Splitting audio files based on silence for long audio file processing or streaming is also possible which brings to question the necessity of these alternatives to MHSA. 

We believe that future work could involve processing large audio files and developing Large Audio Foundation Models, similar to research trends to include more context in Large Language Models (LLM). With this trend in mind, one could imagine doing common LLM tasks, such as summarization, that require long context directly from audio without the need of a transcription.

Nevertheless, as a result of our findings, we believe that unless specifically working with long audio files or from the raw waveform, further speed and memory gains will not be obtained by replacing MHSA with alternatives with linear complexity. We believe that more efficient SSL for speech models might be reached by other architectural changes, pruning/quantization methods, and data selection.

\newpage
\bibliographystyle{IEEEbib}
\bibliography{mybib}

\end{document}